\documentclass[aoas,MSNbibl,nameyear,seceqn,dvips]{arximspdf}
\usepackage{graphicx}

\doi{10.1214/12-AOAS597} 
\volume{7}
\issue{1}
\pubyear{2013}
\firstpage{523}
\lastpage{542}

\makeatletter

\newcommand{\rright}{\right}
\newcommand{\lleft}{\left}
\renewcommand{\citep}[1]{(\citeauthor{#1} \citeyear{#1})}
\newcommand{\citepp}[1]{[\citeauthor{#1} (\citeyear{#1})]}

\def\rank{\operatorname{rank}}
\def\pen{\operatorname{Pen}}

\def\textcolor#1#2{#2}
\makeatother

\begin{document}
\begin{frontmatter}

\title{Joint and individual variation explained (JIVE) for integrated
analysis of multiple data types\thanksref{T1}}
\runtitle{Joint and individual variation explained}
\thankstext{T1}{Supported in part by NIH Grant R01 MH090936-01, NSF
Grant DMS-09-07177, NSF Grant DMS-08-54908 and NIH Grant U24-CA143848.}

\begin{aug}
\author[a]{\fnms{Eric F.} \snm{Lock}\corref{}\ead[label=e1]{lock@email.unc.edu}},
\author[b]{\fnms{Katherine A.} \snm{Hoadley}\ead[label=e2]{hoadley@med.unc.edu}},
\author[a]{\fnms{J.~S.}~\snm{Marron}\ead[label=e3]{marron@email.unc.edu}}
\and
\author[a]{\fnms{Andrew B.} \snm{Nobel}\ead[label=e4]{nobel@email.unc.edu}}
\runauthor{Lock, Hoadley, Marron and Nobel}
\affiliation{University of North Carolina at Chapel Hill}
\address[a]{E.~F. Lock\\
J.~S. Marron\\
A.~B. Nobel\\
Department of Statistics\\
\quad and Operations Research\\
University of North Carolina\\
\quad at Chapel Hill\\
Chapel Hill, North Carolina 27599\\
USA\\
\printead{e1}\\
\phantom{E-mail:\ }\printead*{e3}\\
\phantom{E-mail:\ }\printead*{e4}}

\address[b]{K.~A. Hoadley \\
Lineberger Comprehensive Cancer Center\\
University of North Carolina\\
\quad at Chapel Hill\\
450 West Dr.\\
Chapel Hill, North Carolina 27599\\
USA\\
\printead{e2}}
\end{aug}

\received{\smonth{11} \syear{2011}}
\revised{\smonth{9} \syear{2012}}

%
\begin{abstract}
Research in several fields now requires the analysis of data sets in
which multiple high-dimensional types of data are available for a
common set of objects. In particular, The Cancer Genome Atlas (TCGA)
includes data from several diverse genomic technologies on the same
cancerous tumor samples. In this paper we introduce Joint and
Individual Variation Explained (JIVE), a general decomposition of
variation for the integrated analysis of such data sets. The
decomposition consists of three terms: a low-rank approximation
capturing joint variation across data types, low-rank approximations
for structured variation individual to each data type, and residual
noise. JIVE quantifies the amount of joint variation between data
types, reduces the dimensionality of the data and provides new
directions for the visual exploration of joint and individual
structures. The proposed method represents an extension of Principal
Component Analysis and has clear advantages over popular two-block
methods such as Canonical Correlation Analysis and Partial Least
Squares. A JIVE analysis of gene expression and miRNA data on
Glioblastoma Multiforme tumor samples reveals gene--miRNA associations
and provides better characterization of tumor types.

Data and software are available at \url{https://genome.unc.edu/jive/}.
\end{abstract}

%

\end{frontmatter}

\section{Introduction} \label{Sec1}

Many fields of scientific research now analyze high-dimensional data,
in which a large number of variables are measured for a given set of
experimental objects. Increasingly, those data include multiple
high-dimensional data sets for a common set of objects. Table~\ref
{tab1} gives very diverse examples of such data objects. In this
context we refer to each data set as a \emph{data type} to indicate
that it comes from a distinct mode of measurement or domain.

\begin{table}
\tabcolsep=3pt
\caption{Examples with multiple high-dimensional data types}\label{tab1}
\begin{tabular*}{\textwidth}{@{\extracolsep{\fill}}lcp{190pt}@{}}
\hline
\textbf{Field} & \textbf{Object} & \multicolumn{1}{c@{}}{\textbf{Data types}} \\
\hline
Computational biology & Tissue samples & Gene expression, microRNA,
genotype, protein abundance/activity \\
Chemometrics & Chemicals & Mass spectra, NMR spectra, atomic
composition \\
Atmospheric sciences & Locations & Temperature, humidity, particle
concentrations over time \\
Internet traffic & Websites & Word frequencies, visitor demographics,
linked pages \\
\hline
\end{tabular*}
\end{table}

The motivation for this article is a particular application to
biological data. In biomedical studies, a number of technologies now
commonly collect diverse information on an organism or tissue sample.
The amount of available biological data from multiple platforms and
technologies is expanding rapidly. The 2011 Online Database collection
of \emph{Nucleic Acids Research} lists 1330 publicly available
databases that measure various aspects of molecular and cell biology
\citepp{Galberin}. Large online databases such as ArrayExpress \citepp
{Parkinson} and the UCSC Genome-browser \citepp{Rhead} often contain
multiple data types collected from a common set of samples. Large-scale
projects like The Human Connectome Project \citepp{Sporns} and The
Cancer Genome Atlas \citepp{TCGA} focus on the integrated analysis of
multiple data types.


Well-established multivariate methods can be used to separately analyze
different data types measured on the same set of objects. However,
individual analyses will not capture the critical associations and
potential causal relationships between data types. Furthermore, each
data type can impart unique and useful information. There is a strong
need for new statistical methods that explore associations between
multiple data types and combine data from multiple sources when making
inference about the objects. This motivates an interesting new area of
statistical research.


\subsection{Data} \label{Data}

We describe an application to data from TCGA, an ongoing collaborative
effort funded by the National Cancer Institute (NCI) and the National
Human Genome Research Institute (NHGRI). A goal of TCGA is to
characterize cancer on a molecular level through the analysis and
integration of multidimensional large scale genomic data. The
integration of information from disparate genomic sources has the
promise to provide a more comprehensive understanding of cancer
genetics and cell biology.

We focus, in particular, on a set of 234 Glioblastoma Multiforme (GBM)
tumor samples. GBM is a common and very fatal form of malignant brain
tumor. However, GBM cases are not homogeneous, and an understanding of
systematic distinctions between the tumor samples may lead to more
targeted therapies. \citet{Verheek} classified the TCGA GBM samples
into four subtypes: Neural, Mesenchymal, Proneural and Classical. These
subtypes have distinct expression characteristics, copy number
alterations and gene mutations. In addition, there were clinical
differences across subtypes in response to aggressive therapy.

Copy number aberrations and somatic mutations, and their relation to
gene expression, have been recognized as important aspects of GBM
biology [see, e.g., \citet{Bredel} and \citet{TCGA}]. However, the role
of microRNA (miRNA) data in GBM biology has not been well studied. In
this article we focus on the integrated analysis of miRNA and gene
expression data. These data are two distinct data types, as they are
measured on different platforms and represent different biological components.

Current biological ideas suggest that miRNAs function primarily as
post-transcriptional regulators of gene expression. Typically, they are
considered negative regulators, decreasing gene expression levels. Many
of the algorithms that predict miRNA targets (\emph{TargetScan}, \emph{miRanda},
\emph{Pictar}, and \emph{RNA22}) have vastly different predicted gene lists \citepp
{Peter}. Therefore, miRNA and gene expression relations are not well
understood. However, recent research suggests that miRNAs may be partly
responsible for the expression of well-known tumor activating genes
(oncogenes) and tumor suppressing genes.

Investigating each data type individually would lose some important
relations, considering the inherent interactions between miRNAs and
gene expression. An integrative, multivariate approach is desired. The
data decomposition proposed in Section~\ref{PropMeth} gives new insight
into the joint and individual variation between the miRNA and gene
expression data. Although this decomposition is unsupervised with
respect to the above GBM subtypes, we further investigate how it leads
to a better characterization of these subtypes.

For each tumor sample, there are measures of intensity for 534 miRNAs
and 23,293 genes (messenger RNA). These data are publicly available
from TCGA. The preprocessed data used for this analysis are available
at \url{https://genome.unc.edu/jive}.

\subsection{Proposed method}\label{PropMeth}

Given the biological relation between gene expression and miRNA, it is
reasonable to expect shared patterns in the two sets of measurements.
We refer to such shared patterns as \emph{joint structure}. We also
expect the gene expression data to have systematic variation that is
unrelated to the miRNAs and vice versa. This \emph{individual
structure} can be the result of technical artifacts, but may also be of
biological interest. For example, miRNA regulation is just one of many
factors that can influence gene expression. This structured individual
variation can interfere with finding the important joint signal, just
as joint structure can obscure the important signal that is individual
to a data type.

To separate joint and individual effects, we introduce a method called
Joint and Individual Variation Explained (JIVE). This exploratory
method decomposes a data set into a sum of three terms: a low-rank
approximation capturing joint structure between data types, low-rank
approximations capturing structure individual to each data type and
residual noise. Analysis of individual structure provides a way to
identify potentially useful information that exists in one data type,
but not others. Accounting for individual structure also allows for
more accurate estimation of what is common between data types. As
illustrated in Section~\ref{sec33}, JIVE can identify joint structure
not found by existing methods. It may be used regardless of whether the
dimension of a data set exceeds the sample size. Furthermore, JIVE is
applicable to data sets with more than two data types and has a simple
algebraic interpretation.

\begin{figure}

\includegraphics{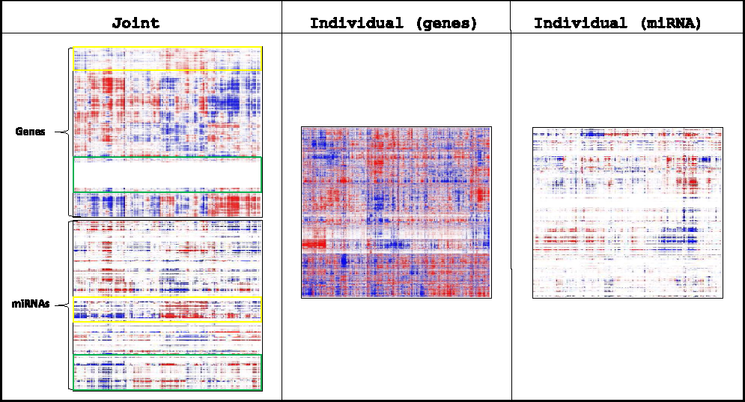}

\caption{JIVE estimates for joint structure and individual structure
in the GBM data. \textcolor{blue}{Blue} corresponds to negative values,
\textcolor{red}{red} positive values. Columns are given in the same
order in the joint structure, and subsets of genes and miRNAs that
share similar patterns are highlighted in \textcolor{green}{green} and
\textcolor{yellow}{yellow}.}
\label{FIntro}
\end{figure}

A heatmap of joint structure and individual structure in the GBM data,
identified by JIVE, is shown in Figure~\ref{FIntro}. Columns
(corresponding to the 234 samples) are shown in the same order for both
gene and miRNA data in the display of joint structure. This common
ordering shows complex structure in both data types, and shared
patterns are present in subsets of genes and miRNAs. JIVE also
identifies a large amount of structured variation individual to the
gene expression data and a lesser degree of individual structure in the
miRNA data. While the individual structure accounts for more
variability in the data than the joint structure, our analysis suggests
that the joint structure is more relevant to the cancer biology.

\section{Model and estimation} \label{Sec2}

\subsection{Formal framework} \label{Sec12}

Formally, we focus on data for multiple matrices $X_1,X_2,\ldots,X_k$
with $k \geq2$. Each matrix has $n$ columns, corresponding to a common
set of $n$ objects. The $i$th matrix $X_i$ has $p_i$ rows, each
corresponding to a variable in a given measurement technology that
varies from matrix to matrix.
For example, in our application to GBM data the rows of $X_1$ contain
gene expression measurements (of dimension $p_1 = 23\mbox{,}293$) and the rows
of $X_2$ contain miRNA measurements (of dimension $p_2 = 534$) for the
same set of 234 tissue samples ($n=234$).
The $k$ matrices may be combined into a single data matrix
\[
X = \lleft[ %
\matrix{ X_1
\vspace*{2pt}\cr
\vdots
\vspace*{2pt}\cr
X_k} %
\rright] \dvtx p \times n,
\]
where $p = p_1+p_2+\cdots+p_k$.\vadjust{\goodbreak}

Direct analysis of $X$ can be problematic as the size and scale of the
constituent data types often differ. To remove baseline differences
between data types, it is helpful to row-center the data by subtracting
the mean within each row. Data types may also be of different dimension
($p_i$) or differ in variability. To circumvent cases where ``the
largest data set wins,'' it is helpful to scale each data type\vspace*{1pt} by its
total variation, or sum-of-squares. In particular, for each $i$ define
$X_i^{\mathrm{scaled}}$ = $\frac{X_i}{\Vert X_i\Vert }$, where $\Vert \cdot\Vert $
defines the Frobenius norm
$\Vert A\Vert^2 = \sum_{i,j} a_{ij}^2$.
Then, $\Vert X_i^{\mathrm{scaled}}\Vert  = 1$ for each~$i$, and each
data type contributes equally to the total variation of the
concatenated matrix
%
\begin{equation}
\label{Scaling} X^{\mathrm{scaled}} = \lleft[ %
\matrix{
X_1^{\mathrm{scaled}} \vspace*{2pt}
\cr
\vdots\vspace*{2pt}
\cr
X_k^{\mathrm{scaled}} } %
\rright].
\end{equation}

\subsection{Model}

Let $X_1, X_2, \ldots, X_k$ be matrices as in Section~\ref{Sec12}, scaled
appropriately. Joint structure is represented by a single $p \times n$
matrix of rank $r < \rank(X)$. Individual structure for each $X_i$ is
represented by a $p_i \times n$ matrix of rank $r_i < \rank(X_i)$.

More formally, let $A_i$ be the matrix representing the individual
structure of $X_i$, and let $J_i$ be the submatrix of the joint
structure matrix that is associated with $X_i$. Then, the unified JIVE
model is
%
\begin{eqnarray}
\label{model} X_1 &=& J_1+A_1 +
\varepsilon_1,
\nonumber
\\
&\vdots&
\\
X_k &=& J_k + A_k + \varepsilon_k,
\nonumber
\end{eqnarray}
where $\varepsilon_i$ are $p_i \times n$ error matrices of independent
entries with $\mathbb{E}(\varepsilon_i) = 0_{p_i \times n}$.
Let
\[
J = \lleft[ %
\matrix{J_1
\vspace*{2pt}\cr
\vdots
\vspace*{2pt}\cr
J_k }
\rright]
\]
denote the joint structure matrix.
The model imposes the rank constraints $\rank(J) = r$ and $\rank(A_i) =
r_i$ for $i = 1,\ldots,k$.
Furthermore, we require that the rows of joint and individual
structures are orthogonal:
$J A_i^T = 0_{p \times p_i}$ for $i=1,\ldots,k$.
Intuitively, this means that sample patterns responsible for joint
structure between data types are unrelated to sample patterns
responsible for individual structure. This requirement does not
constrain the model, in that any matrix in the form (\ref{model}) can
be written equivalently with orthogonality between joint and individual
structures. Furthermore, the orthogonality constraint assures that the
joint and individual components in (\ref{model}) are uniquely
determined (see the supplementary material [\citet{Lock}] for more details). It is remarkable that no
further orthogonality constraints (say, between the column space of
$J_i$ and the column space of $A_i$, or between the row spaces of each
$A_i$) are required to make the decomposition identifiable.

\subsection{Estimation} \label{est}
Here we discuss estimation of the model for fixed ranks
$r,r_1,\ldots,r_k$. The choice of these ranks is important to accurately
quantify the amount of joint and individual structures. The supplementary material [\citet{Lock}]
describes a permutation testing approach to rank selection.

Joint and individual structures are estimated by minimizing the sum of
squared error. Let $R$ be the $p \times n$ matrix of residuals after
accounting for joint and individual structures:
\[
R = \lleft[ %
\matrix{ R_1
\vspace*{2pt}\cr
\vdots
\vspace*{2pt}\cr
R_k }
\rright] = \lleft[ %
\matrix{ X_1 -J_1-A_1
\vspace*{2pt}\cr
\vdots
\vspace*{2pt}\cr
X_k -J_k-A_k }
\rright]
.
\]
We estimate the matrices $J$ and $A_1,\ldots, A_k$ by minimizing
$\Vert R\Vert^2$ under the given ranks. This is accomplished by iteratively
estimating joint and individual structures:
\begin{itemize}
\item Given $J$, find $A_1,\ldots,A_k$ to minimize $\Vert R\Vert $.
\item Given $A_1,\ldots,A_k$, find $J$ to minimize $\Vert R\Vert $.
\item Repeat until convergence.
\end{itemize}

The joint structure $J$ minimizing $\Vert R\Vert $ is equal to the first $r$
terms in the singular value decomposition (SVD) of X with individual
structure removed. The estimated individual structure for $X_i$ is
equal to the first $r_i$ terms of the SVD of $X_i$ with the joint
structure removed. The estimate of individual structure for $X_i$ will
not change those for $X_j$, $j \neq i$ and, hence, the $k$ individual
approximations minimize $\Vert R\Vert $ for fixed joint structure. Pseudocode
for this iterative algorithm is given in the supplementary material [\citet{Lock}]. Computing time
can be improved by reducing the dimensionality of $X_1,\ldots,X_k$ at
the outset via their SVD (see the supplementary material [\citet{Lock}]).

\begin{figure}

\includegraphics{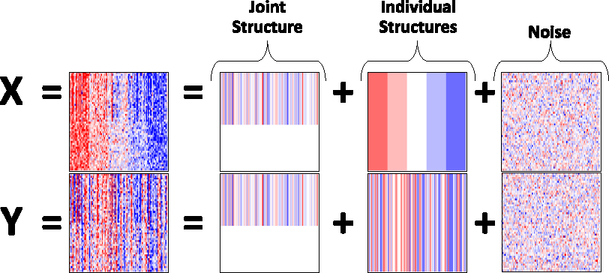}

\caption{$X$ and $Y$ are generated by adding together joint structure,
individual structure and noise. \textcolor{blue}{Blue} corresponds to
negative values, \textcolor{red}{red} positive values.}
\label{FDiagram2}
\end{figure}

The iterative method is monotone in the sense that $\Vert R\Vert $ decreases at
each step. Thus, $\Vert R\Vert $ converges to a coordinate-wise minimum, which
can not be improved by changing the estimated joint or individual,
structure. Further convergence properties of the algorithm are
currently under study.

\subsection{Illustrative example} \label{Sec24}

As a basic illustration we generate two matrices, $X$~and $Y$, with
simple patterns corresponding to joint and individual structures. The
simulated data are depicted in Figure~\ref{FDiagram2}. Both $X$ and
$Y$ are of dimension $50 \times100$, that is, each has 50 variables
measured for the same 100 objects. A common pattern $V$ of 100
independent standard normal variables is added to half of the rows in
$X$ and half of the rows in $Y$. This represents the joint structure
between the two data sets. Structure individual to $X$ is generated by
partitioning the objects into five groups, each of size twenty. Those
columns corresponding to group 1, 2, 3, 4 or 5 have $-2$, $-1$, 0, 1, 2
added to each row of $X$, respectively. Structure individual to $Y$ is
generated similarly, but the groups are randomly determined and are
therefore independent of the groups in $X$. Finally, independent
$N(0,1)$ noise is added to both $X$ and $Y$. Note that the important
joint structure is visually obscured.

The common pattern $V$ represents an underlying phenomenon that
contributes to several variables in both $X$ and $Y$. Practically, the
individual structure in $X$ (or~$Y$) may correspond to an experimental
grouping of the measured variables in~$X$ ($Y$) not present in $Y$
($X$), for example, \emph{batch} effects in microarray data. Our goal
is to identify both the common underlying phenomenon and individual
group effects.

\begin{figure}

\includegraphics{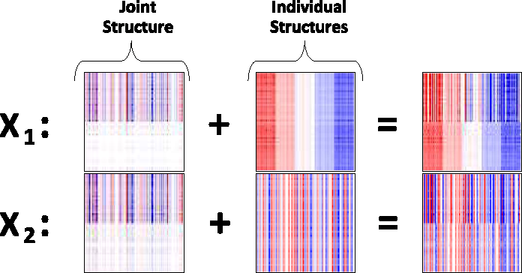}

\caption{JIVE estimates for joint structure and individual structure.
\textcolor{blue}{Blue} corresponds to negative values, \textcolor{red}{red} positive values.}
\label{FDiagramJive}
\end{figure}

Figure~\ref{FDiagramJive} shows the JIVE estimate for joint structure,
JIVE estimates for individual structure and the fit given by the sum of
joint and individual structures. Estimates closely resemble the true
signal in Figure~\ref{FDiagram2}.

\subsection{GBM data} \label{GBMest}

\begin{figure}

\includegraphics{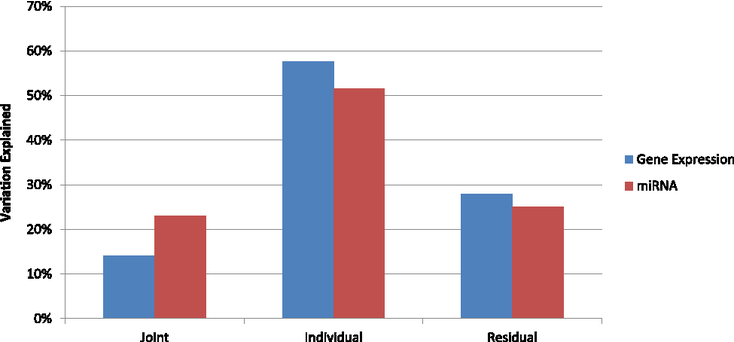}

\caption{Percentage of variation (sum of squares) explained by
estimated joint structure, individual structure and residual noise for
miRNA and gene expression data.}
\label{FPieCharts}
\end{figure}

As the gene expression and miRNA data for the GBM samples differ in
dimension and variability, they were scaled as in Section~\ref{Sec12}.
Permutation testing (see the supplementary material [\citet{Lock}]) was used to determine the ranks
of estimated joint and individual structures. The test (using $\alpha=
0.01$, and 1000 permutations) identified:
\begin{itemize}
\item rank 5 joint structure
\item rank 33 structure individual to gene expression
\item rank 13 structure individual to miRNA.
\end{itemize}
The percentage of variation (sum of squares) explained in each data set
by joint structure, individual structure and residual noise is shown in
Figure~\ref{FPieCharts}. This illustrates how the JIVE decomposition
can be used to quantify and compare the amount of shared and individual
variation between data types. As shown in Figure~\ref{FPieCharts},
joint structure is responsible for more variation in miRNA than in gene
expression (23\% and 14\%, resp.), and the gene expression data
has a considerable amount of structured variation (58\%) that is
unrelated to miRNA. This is consistent with current biological
understanding, as miRNAs are just one of several factors that can
influence gene expression.\looseness=-1

\begin{figure}[b]

\includegraphics{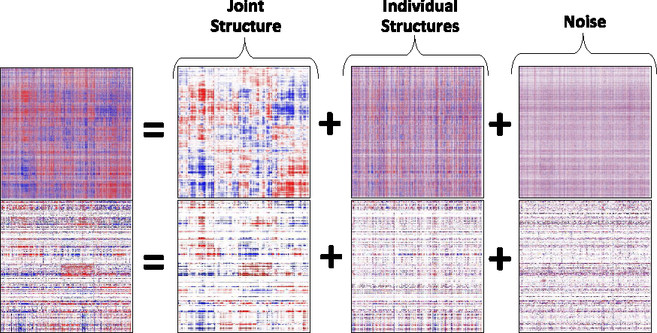}

\caption{Heatmaps of low-rank estimates for joint structure,
individual structure and residual noise in the gene expression (top)
and miRNA (bottom) data. \textcolor{blue}{Blue} corresponds to negative
values, \textcolor{red}{red} positive values. Columns have the same
order in all heatmaps.}
\label{FTCGAheatmaps}
\end{figure}

Heatmaps of the low-rank estimates for joint structure, individual
structure and residual noise\vadjust{\goodbreak} are shown in Figure~\ref{FTCGAheatmaps}.
Columns have the same order in all heatmaps. This reveals the shared
patterns present in the joint structure from Figure~\ref{FIntro}, but
little of the structure that is present in the individual estimates.
%

\section{Model factorization} \label{ModelFactorization}

\subsection{Relation to PCA}\label{sec32}

The JIVE model can be factorized as in \emph{Principal Component
Analysis} (PCA). For a row-centered\vadjust{\goodbreak} $p \times n$ matrix $X$, the first
$r$ principal components yield the rank $r$ approximation
\[
X \approx U S,
\]
where $S (r \times n)$ contains the sample scores and $U (p
\times r)$ contains the variable loadings for the first $r$
components.

As in PCA, the rank $r$ joint structure matrix $J$ in the JIVE model
can be written as $U S$, where $U$ is a $p \times r$ loading matrix and
$S$ is an $r \times n$ score matrix. Let
\[
U = \lleft[ %
\matrix{ U_1
\vspace*{2pt}\cr
\vdots
\vspace*{2pt}\cr
U_k }
\rright],
\]
where $U_i$ gives the loadings of the joint structure
corresponding to the rows of~$X_i$. The rank $r_i$ individual structure
matrix $A_i$ for $X_i$ can be written as $W_i S_i$, where $W_i$ is a
$p_i \times r_i$ loading matrix and $S_i$ is an $r_i \times n$ score
matrix. Then, the low-rank decomposition of $X_i$ into joint and
individual structures is given by $X_i \approx U_i S + W_i S_i.$
This gives the factorized model
%
\begin{eqnarray}
\label{modelPCA} X_1 &=& U_1 S+W_1
S_1 + R_1 ,
\nonumber\\
&\vdots&
\\
X_k &=& U_k S +W_k S_k
+R_k.
\nonumber
\end{eqnarray}
Joint structure is represented by the common score matrix $S$. These
scores summarize patterns in the samples that explain variability
across multiple data types. The loading matrices $U_i$ indicate how
these joint scores are expressed in the rows (variables) of data type
$i$. The score matrices $S_i$ summarize sample patterns individual to
data type $i$, with variable loadings~$W_i$.

\subsection{GBM data}

\begin{figure}

\includegraphics{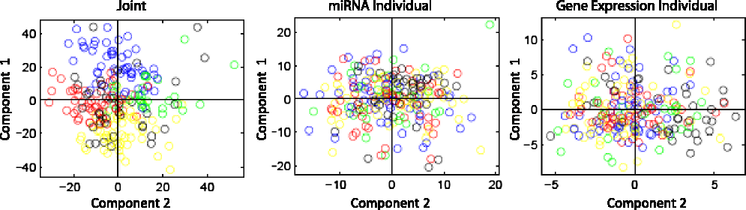}

\caption{Scatterplots of sample scores for the first two joint
components, first two individual miRNA components and first two
individual gene expression components. Samples are colored by subtype:
Mesenchymal (\textcolor{yellow}{yellow}), Proneural (\textcolor
{blue}{blue}), Neural (\textcolor{green}{green}) and Classical
(\textcolor{red}{red}). Samples colored black were assayed after the
initial subtype analysis and are considered unclassified.}
\label{FScatterPlots}
\end{figure}

Sample scores for joint structure, matrix $S$ in equation~(\ref
{modelPCA}), reveal sample patterns that are present across the miRNA
and gene expression data. Sample scores for individual structure,
matrices $S_1$ and $S_2$ in equation~(\ref{modelPCA}), reveal sample
patterns that are individual to each data type. Figure~\ref
{FScatterPlots} shows separate scatterplots of the sample scores for
the first two principal components of estimated joint structure, the
first two components individual to miRNA, and the first two components
individual to gene expression. Subtype distinctions are clearly present
in the scatterplot of joint scores, but a subtype effect is not
visually apparent in either of the individual scatterplots.

\begin{table}[b]
\caption{SWISS scores for TCGA subtypes. Lower scores indicate more
subtype distinction}\label{SWISS}
\begin{tabular*}{\textwidth}{@{\extracolsep{\fill}}lcc@{}}
\hline
{Data}
&Gene expression & 0.8431 \\
&miRNA & 0.8763 \\[3pt]
{JIVE}
&Joint & 0.7678 \\
&Gene expression individual & 0.9019 \\
&miRNA individual & 0.9284 \\ \hline
\end{tabular*}
\end{table}

Since the subtypes are defined by gene expression clustering, their
appearance in Figure \ref{FScatterPlots} is not surprising. However,
the clustering apparent in the joint plot shows involvement of miRNA in
the differentiation of these subtypes. It is interesting that a subtype
effect is not apparent in either scatterplot for individual structure,
suggesting that this variation is driven by other biological
components. This is remarkable, as the fraction of gene expression
variation explained by joint structure (see Figure~\ref{FPieCharts})
is small.

To numerically compare the extent to which subtype distinctions are
present, we consider the standardized within-subtype sum of squares
\[
\operatorname{SWISS}(A) = \frac{\sum_i \sum_j (A_{ij}-\bar{A}_{i,s(j)})^2}{\sum_i \sum_j (A_{ij}-\bar{A}_{i \cdot})^2},
\]
where $s(j) = \{k \dvtx \mathrm{samples}\ j \mbox{ and } k \mbox{ belong to
the same subtype}\}$. This represents the variability within subtypes
(across all rows) as a proportion of total variability. Table \ref
{SWISS} gives SWISS scores for the gene expression and miRNA data, and
SWISS scores for the JIVE estimates of joint and individual structures.
A permutation test described in \citet{Cabanski} concludes that the
four subtypes are significantly more distinguished on the estimated
joint structure than on the gene expression and miRNA data ($p <
0.001$; $10\mbox{,}000$ permutations). SWISS scores for individual structure
in gene expression and miRNA are close to one, as subtype distinctions
are almost entirely represented in the joint structure between the two
data types. This suggests that miRNA may play a greater role in GBM
biology than previously thought.

In general, these analyses illustrate how an unsupervised, integrated
analysis across multiple data types can result in a better distinction
between subtypes or other biological classes. One could conduct a
similar analysis to investigate how the JIVE components relate to
survival or other clinical factors, rather than subtype. Furthermore, a
direct cluster analysis on the JIVE components could be used to
identify sample groups that are distinguished across multiple data types.

\section{Comparison with existing methods}\label{Sec3}

\subsection{Existing methods} \label{Sec31}

One approach to the analysis of multiple data sets is to mine the data
for variable-by-variable associations. In computational biology,
large-scale correlation studies can identify millions of pairwise
variable associations between genomic data types [see, e.g.,
\citet{Gilad}]. Furthermore, network models can link associated
variables across and within data types [see \citet{Adourian}]. However,
analysis of variable-by-variable associations alone does not identify
the global modes of variation that drive associations across and within
data types, which is the focus of this paper.

PCA of the block-scaled matrix $X^{\mathrm{scaled}}$ in (\ref{Scaling})
coincides with \emph{Consensus PCA} [\citet{Wold}, \citet{Westerhuis}]. This
direct approach is also used by the iCluster method~[Shen, Olshen and Ladanyi (\citeyear{icluster})],
which performs clustering based on a factor analysis of the
concatenated matrix $X$. While these methods synthesize information
from multiple data types, they do not distinguish between joint or
individual effects.

\emph{Canonical Correlation Analysis} (CCA) \citepp{Hotelling} is a
popular method to globally examine the relation between two sets of
variables. If $X_1$ and $X_2$ are two data sets on a common set of
samples, the first pair of canonical loadings (variable weights) $u_1$
and $u_2$ are unit vectors maximizing $\operatorname{Corr}(u_1^T X_1, u_2^T X_2)$.
Geometrically, $u_1$ and $u_2$ can be interpreted as the pair of
directions that maximize the correlation between $X_1$ and $X_2$.
Sample projections on the canonical loadings, $u_1^T X_1$ and $u_2^T
X_2$, give the canonical \emph{scores} for $X_1$ and $X_2$. Subsequent
CCA directions can be found by enforcing orthogonality with previous
directions. For data sets with $p_1>n$ or $p_2>n$ the CCA directions
are not well defined, and overfitting is often a problem even when
$p_1,p_2 <n$. Hence, standard CCA is typically not applicable to
high-dimensional data.

\emph{Partial Least Squares} (PLS) \citepp{Hwold} directions are defined
similarly to CCA, but maximize covariance rather than correlation. PLS
is appropriate for high-dimensional data. However, \citet{Trygg}
examine how structured variation in $X_1$ not associated with $X_2$
(and vice versa) can drastically alter PLS scores, making the
interpretation of such scores problematic. Their solution, called
O2-PLS, seeks to remove structured variation in $X_1$ not linearly
related to $X_2$ (and vice versa) from the PLS components. As such,
O2-PLS components are often more representative of the true joint
structure between two data types. However, the restriction of O2-PLS
(and PLS) to pairwise comparisons limits their utilty in finding common
structure among more than two data types.

\citet{Witten} recently introduced \emph{Multiple Canonical
Correlation Analysis} (mCCA) to explore associations and common
structure on two or more data sets. For $X_1, X_2, \ldots, X_k$ as in
Section \ref{Sec12}, standardized so that each row has mean 0 and
standard deviation 1, the standard mCCA loading vectors
$u_1,u_2,\ldots,u_k$ satisfy
\[
\mathop{\arg\max}_{\Vert u_1\Vert =\cdots=\Vert u_k\Vert  = 1} \sum_{i<j}
u_i^T X_i X_j^T
u_j = \sum_{i<j} \operatorname{Cov}
\bigl(u_i^T X_i, u_j^T
X_j\bigr) .
\]

As such, mCCA can be viewed as a natural extension of PLS to
more than two data types.

\citet{Di} develop multi-level functional PCA (MF-PCA) for the analysis
of variation between and within grouped samples of functional data.
Similar in spirit to JIVE, MF-PCA yields a sum of two PCA
decompositions: one for variability between groups and one for
variability within groups. We stress that JIVE is designed for analysis
across disparate data types, while MF-PCA analyzes grouped observations
on the same functional data type. More substantively, MF-PCA does not
model shared structure and individual structure. Rather, MF-PCA models
structure among main effects (e.g., at the group level) and structured
variation about those main effects (e.g., at the sample level), in the
context of functional data.

\subsection{Illustrative example}\label{sec33}

\begin{figure}

\includegraphics{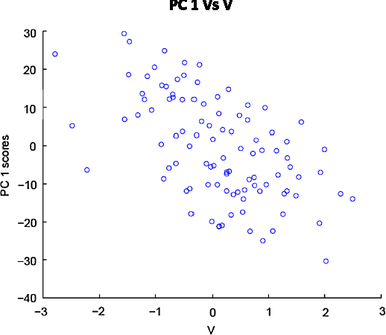}

\caption{Scatterplot of the first consensus principal component scores
vs the joint signal~$V$. The scores are weakly associated with the
joint signal.}
\label{FDiagram3}
\end{figure}

We return to the example introduced in Section~\ref{Sec24}. Consensus
PCA of the concatenated matrix $[{X' \enskip   Y'}]'$ does a poor job of
finding the joint structure. The scatterplot in Figure~\ref{FDiagram3}
shows a weak association between the first principal component scores
and the joint response $V$. This is because PCA of the concatenated
data is driven by all variation in the data, joint or individual.

\begin{figure}[b]

\includegraphics{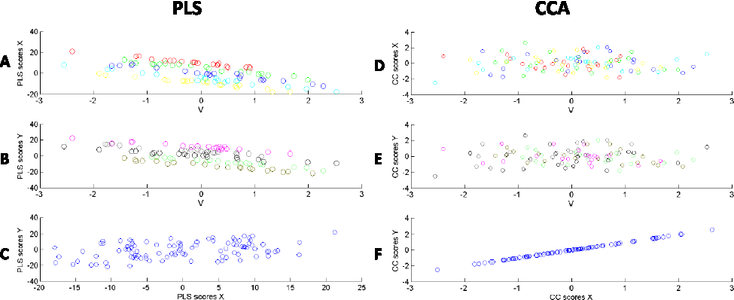}

\caption{Scores for the first PLS, and CCA, directions for $X$ and
$Y$. Panels \textup{(A)} and \textup{(B)} show some association between the PLS scores
and the joint signal $V$. Points are colored by simulated group in both
$X$ and $Y$, and are more highly associated with $V$ within each group.
The two PLS directions have a weak correlation \textup{(C)}. The CCA directions
correlate with each other \textup{(F)}, but not the common signal $V$ (\textup{D} and \textup{E}).
This illustrates the tendency of CCA to overfit.}
\label{FDiagram4}
\end{figure}

Figure~\ref{FDiagram4} shows an analysis of PLS and CCA for $X$ and
$Y$. The scores for the first PLS direction for $X$ and $Y$ show a weak
association (panel C). Furthermore, the PLS scores are not strongly
related to the joint response~$V$ (panels A and B). Scores for $X$ and
$Y$ show a stronger association with $V$ within groups, indicating how
for PLS individual structure can interfere with the identification of
joint structure. The CCA scores are very highly correlated (panel F),
but show nearly no association with the joint or individual structures
(panels D and E). This illustrates the tendency of CCA to overfit on
high-dimensional data.

\begin{figure}

\includegraphics{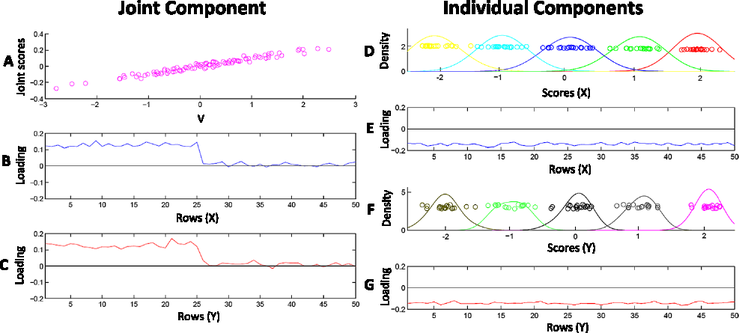}

\caption{Scores and loadings for joint and individual components in
the JIVE decomposition. Joint scores are highly associated with the
common signal $V$ (panel \textup{A}). Individual scores distinguish groups
specific to $X$ and $Y$ (\textup{D} and \textup{F}). Joint loadings (\textup{B} and \textup{C}) show a
strong effect (difference from zero) on half of the variables in $X$
and $Y$. Individual loadings (\textup{E} and \textup{G}) show a similar effect on all
variables in $X$ and $Y$.}
\label{FDiagram5}
\end{figure}

Next we consider the JIVE analysis of $X$ and $Y$. Scores and loadings
for the joint component and both individual components are shown in
Figure~\ref{FDiagram5}. JIVE is able to find the true joint signal
between the two data sets, as joint scores are closely associated with
the common response $V$ (panel A). Furthermore, individual scores do a
good job of distinguishing the groups specific to $X$ and $Y$ (panels D
and~F). The joint signal was added to only the first 25 variables in
$X$ and $Y$, which is reflected in the joint loadings (panels B and C).
The individual groups were defined on all 50 variables for both $X$ and
$Y$, which is reflected in the individual loadings (panels E and G).
Note that joint and individual loadings are not constrained to be
orthogonal, which gives the analysis more flexibility.

\section{Variable sparsity} \label{Sec4}
In many practical applications, important structure between samples or
objects is present on only a small subset of the measured variables.
This motivates the use of sparse methods, in which only a subset of
variables contribute to a fitted model. Sparse versions of exploratory
methods such as PCA \citepp{ShenHuang}, PLS \citepp{LeCao} and CCA \citepp
{Parkhomenko} already exist.

Here, we describe the use of a penalty term to induce variable sparsity
in the JIVE decomposition. Sparsity is accomplished if some of the
variable loadings for joint and individual structures [$U$ and $W_i$ in
equation~(\ref{modelPCA})] are exactly 0. For weights $\lambda$ and
$\lambda_i$, we minimize the penalized sum of squares
\[
\Vert R\Vert^2 +\lambda \pen(U) +{\sum}_{i}
\lambda_i \pen(W_i),
\]
where $\pen$ is a penalty designed to induce sparsity in the loading
vectors. In our implementation, $\pen$ is an $L1$ penalty analogous to
Lasso regression \citepp{Tibshirani}, namely,
\[
\pen(A) ={\sum}_{i,j} |a_{ij}| .
\]
Under this penalty, loadings of variables with a small or insignificant
contribution tend to shrink to 0. Other sparsity-inducing penalties
(e.g., hard thresholding) may be substituted for $L1$ penalization.

\begin{figure}[b]

\includegraphics{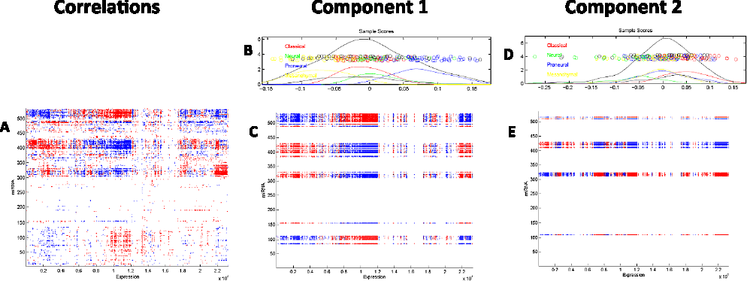}

\caption{Plot of gene--miRNA correlations \textup{(A)}, and scores and loadings
for the first two sparse joint components \textup{(B--E)}. In \textup{(A)}, gene--miRNA
pairs are colored \textcolor{red}{red} if they have a significant
positive correlation and \textcolor{blue}{blue} if they have a
signficant negative correlation ($P<10^{-5}$). Panels \textup{(B)} and \textup{(D)} show
sample scores for the first two joint components, colored by subtype.
Panels \textup{(C)} and \textup{(E)} display gene--miRNA pairs where both have nonzero
loadings. Pairs are colored \textcolor{red}{red} if both gene and miRNA
loadings have the same sign, \textcolor{blue}{blue} otherwise. In
panels \textup{(A)}, \textup{(C)} and \textup{(E)} genes and miRNAs are ordered separately by
average linkage correlation clustering.}
\label{FSparsePlots}
\end{figure}

Estimation with sparsity is accomplished by an iterative procedure
analogous to that
used for the nonsparse case:
\begin{itemize}
\item Given $J$, find $A_i$ to minimize $\Vert R_i\Vert^2+\lambda_i   \pen
(W_i)$ for each $i=1,\ldots,k$.
\item Given $A_1,\ldots,A_k$, find $J$ to minimize $\Vert R\Vert^2+\lambda  \pen(U)$.
\item Repeat until convergence.
\end{itemize}

At each iteration, the sparsity penalty is incorporated
through the use of a sparse singular value decomposition (SSVD),
adapted from \citet{Lee}. The weights $\lambda$, $\lambda_i$ may be
pre-specified or estimated via the \emph{Bayesian Information
Criterion} (BIC) \citepp{Schwarz} at each iteration.

Inducing sparsity in the joint structure effectively identifies subsets
of variables within each data type that are associated. Examination of
the joint sample scores, in turn, reveals sample patterns that drive
these associations.

\subsection{GBM data} \label{Sec41}

A natural way to explore associations between individual
genes and miRNAs is to compute the matrix of all gene--miRNA
correlations, and then examine the set of significant
correlations. Panel A of Figure~\ref{FSparsePlots} shows a heatmap of
the significant gene--miRNA correlations.

A sparse implementation of JIVE provides an alternative approach to identifying
gene--miRNA associations, and can reveal additional structure. Panel B
shows the sample scores in the first joint component resulting from a
sparse JIVE analysis of the data. Panel C shows all the gene--miRNA
pairs with the property that both that gene and miRNA
have nonzero loadings in the first joint component.
Thus, the nonzero entries of the heatmap have the form of
a Cartesian product. We note that the nonzero entries
in panel C closely match those in the
correlation map of panel~A, and that the signs of these
entries also show good agreement. Scores for the first joint component
(panel B) distinguish the Mesenchymal and Proneural subtypes, suggesting
that differences between these sample groups are driving
the first joint component, and appear to influence the correlation
structure of the data as well.

\begin{figure}[b]

\includegraphics{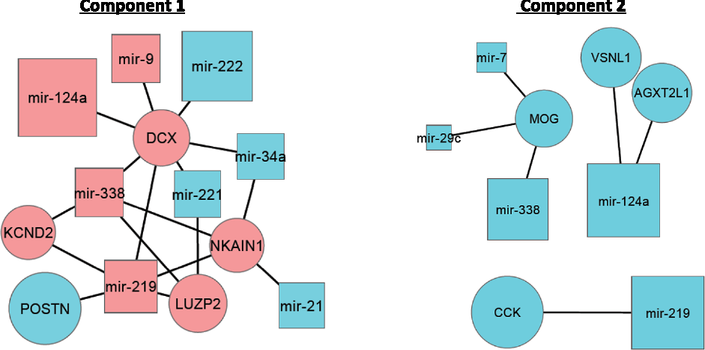}

\caption{Network of predicted gene--miRNA interactions for the first
two sparse joint components. Gene-miRNA pairs are linked if the miRNA
is predicted to target the gene in two or more of the four databases
\emph{miRanda}, \emph{Pictar}, \emph{RNA22} and \emph{TargetScan}.
Genes are shown as circles, and miRNAs are shown as squares. All
predicted targets are shown between the 10 miRNAs with the largest
absolute loading and the 10 genes with the largest absolute loading,
for both components (genes and miRNAs with no targets are not shown).
Genes and miRNAs with a positive loading are colored red, and with a
negative loading are colored blue. The icon size is proportional to the
absolute loading.}
\label{FNetworks}
\end{figure}

Panels D and E display sample scores and nonzero loadings for the
second joint component. Panel D shows that the second joint component
distinguishes
the Neural and Classical subtypes. We note that panel E is
markedly different from panel~A, indicating that the second
joint component is capturing associations between the
expression of genes and miRNA that are not immediately
apparent from the consideration of correlations alone.
Indeed, these associations appear to be masked by variation captured in
the first joint component.

Of primary interest are the biological relations between the genes and
miRNAs. Figure~\ref{FNetworks} displays a network of possible
gene--miRNA interactions for each of the first two joint components,
constructed from genes and miRNAs with large absolute loadings. A
gene--miRNA pair is linked if the miRNA is predicted to regulate the
expression of the gene, based on its DNA sequence. In particular, we
use the mirWalk target prediction module \citepp{Dweep} and include
those pairs that are given in two or more of the miRNA target databases
\emph{miRanda}, \emph{Pictar}, \emph{RNA22} and \emph{TargetScan}. We
caution that current methods to predict miRNA targets are inexact, and
linked gene--miRNA pairs only indicate potential causal relations.
Nevertheless, each joint component includes several predicted
gene--miRNA interactions.

We further examine the individual genes and miRNAs that contribute the
most to the first two joint components. The POSTN gene, which has the
largest loading among genes in the first joint component, encodes the
protein Periostine. Over-expression of Periostine is frequently
reported in cancerous tumor cells and is suspected to facilitate cell
motility (the ability of a cancer cell to migrate quickly and
spontaneously) \citepp{Gillan}. In GBM, the downregulation of POSTN
expression by mir-219 has recently been linked to differences in
survival and time to disease progression \citepp{Zinn}. Interestingly,
the miRNA mir-124a has the largest loading in both components, and a
recent study suggests that this miRNA may also play an important role
in GBM cell motility \citepp{Fowler}.

\section{Summary and discussion} \label{Sec5}

TCGA and similar projects are providing researchers with access to an
increasing number of data sets that consist of multiple
high-dimensional data types. However, there are relatively few general
statistical methods for the analysis of such integrated data sets, and
the unique features of JIVE provide a powerful new approach. JIVE finds
both the coordinated activities of multiple data types, as well as
those features unique to a particular data type. We demonstrate how
accounting for joint structure can lead to better estimation of
individual structure and vice versa. Our application of JIVE to gene
and miRNA data on GBM tumor samples has provided better
characterization of tumor types and better understanding of the
biological interactions between the given data types.

JIVE does not require that the data are ordered, as in \emph{functional
data analysis}. Regularization techniques such as smoothing may improve
the method for functional data. As pointed out by an Associate Editor,
JIVE estimates for joint and individual structures are not robust to
outliers. Exploratory methods such as PCA suffer similarly, and robust
versions of PCA have recently been developed [see, e.g., \citet
{Candes}]. A robust version of JIVE is an interesting potential
extension. Further, any missing values must be imputed prior to
computing JIVE estimates. An approach that explicitly accounts for
missing values in the estimation of joint and individual structures is
another potential extension.

The statistical properties of the algorithm also deserve further
attention. In particular, measures of confidence (e.g., for variation
explained by joint and individual structures) would be useful. Standard
resampling techniques such as bootstrapping may help in this regard.
However, factors such as the discrete nature of the ranks must be
considered carefully.

While this paper focuses on vertically integrated biomedical data, the
JIVE model and algorithm are very general and may be useful in other
contexts. A similar approach can be applied to horizontally integrated
data, in which disparate sets of samples (e.g., sick and healthy
patients) are available on the same data type. In finance, JIVE has the
potential to improve on current models that explain variation across
and within disparate markets [see \citet{Bekaert}]. These applications
are currently under study.

\section*{Acknowledgments}
We wish to thank the Editors and referees for their constructive
comments. We especially appreciate the diligent work of the Associate
Editor on this article.

\begin{supplement}[id=suppA]
\stitle{Additional Material}
\slink[doi]{10.1214/12-AOAS597SUPP}  
\sdatatype{.pdf}
\sfilename{aoas597\_supp.pdf}
\sdescription{The supplementary article~\citet{Lock} provides
additional details and further validation of the JIVE method. This includes:
\begin{itemize}
\item A proof concerning the existence and uniqueness of the decomposition.
\item A description of the permutation approach to rank selection.
\item Pseudocode for the algorithm.
\item A discussion of computing time and efficiency.
\item A discussion of invariance properties.
\item Results from the application of JIVE to many diverse simulated
data sets.
\end{itemize}}
\end{supplement}

%

\printaddresses


\begin{thebibliography}{33}

\bibitem[\protect\citeauthoryear{Adourian et~al.}{2008}]{Adourian}
\begin{barticle}[author]
\bauthor{\bsnm{Adourian},~\bfnm{A.}\binits{A.}},
  \bauthor{\bsnm{Jennings},~\bfnm{E.}\binits{E.}},
  \bauthor{\bsnm{Balasubramanian},~\bfnm{R.}\binits{R.}},
  \bauthor{\bsnm{Hines},~\bfnm{W.}\binits{W.}},
  \bauthor{\bsnm{Damian},~\bfnm{D.}\binits{D.}},
  \bauthor{\bsnm{Plasterer},~\bfnm{T.}\binits{T.}},
  \bauthor{\bsnm{Clish},~\bfnm{C.}\binits{C.}},
  \bauthor{\bsnm{Stroobant},~\bfnm{P.}\binits{P.}},
  \bauthor{\bsnm{McBurney},~\bfnm{R.}\binits{R.}},
  \bauthor{\bsnm{Verheij},~\bfnm{E.}\binits{E.}},
  \bauthor{\bsnm{Bobeldijk},~\bfnm{I.}\binits{I.}},
  \bauthor{\bsnm{Greef},~\bfnm{J.}\binits{J.}},
  \bauthor{\bsnm{Lindberg},~\bfnm{J.}\binits{J.}},
  \bauthor{\bsnm{Kenne},~\bfnm{K.}\binits{K.}},
  \bauthor{\bsnm{Andersson},~\bfnm{U.}\binits{U.}},
  \bauthor{\bsnm{Hellmold},~\bfnm{H.}\binits{H.}},
  \bauthor{\bsnm{Nilsson},~\bfnm{K.}\binits{K.}},
  \bauthor{\bsnm{Salterd},~\bfnm{H.}\binits{H.}} \AND
  \bauthor{\bsnm{Schuppe-Koistinenc},~\bfnm{I.}\binits{I.}}
(\byear{2008}).
\btitle{Correlation network analysis for data integration and biomarker
  selection.}
\bjournal{Molecular BioSystems}
\bvolume{4}
\bpages{249--259}.
\bptok{imsref}%
\end{barticle}
\endbibitem

\bibitem[\protect\citeauthoryear{Bekaert, Hodrick and Zhang}{2009}]{Bekaert}
\begin{barticle}[author]
\bauthor{\bsnm{Bekaert},~\bfnm{G.}\binits{G.}},
  \bauthor{\bsnm{Hodrick},~\bfnm{R.}\binits{R.}} \AND
  \bauthor{\bsnm{Zhang},~\bfnm{X.}\binits{X.}}
(\byear{2009}).
\btitle{International stock return comovements.}
\bjournal{J.~Finance}
\bvolume{64}
\bpages{2591--2626}.
\bptok{imsref}%
\end{barticle}
\endbibitem

\bibitem[\protect\citeauthoryear{Bredel et~al.}{2009}]{Bredel}
\begin{barticle}[pbm]
\bauthor{\bsnm{Bredel},~\bfnm{Markus}\binits{M.}},
  \bauthor{\bsnm{Scholtens},~\bfnm{Denise~M.}\binits{D.~M.}},
  \bauthor{\bsnm{Harsh},~\bfnm{Griffith~R.}\binits{G.~R.}},
  \bauthor{\bsnm{Bredel},~\bfnm{Claudia}\binits{C.}},
  \bauthor{\bsnm{Chandler},~\bfnm{James~P.}\binits{J.~P.}},
  \bauthor{\bsnm{Renfrow},~\bfnm{Jaclyn~J.}\binits{J.~J.}},
  \bauthor{\bsnm{Yadav},~\bfnm{Ajay~K.}\binits{A.~K.}},
  \bauthor{\bsnm{Vogel},~\bfnm{Hannes}\binits{H.}},
  \bauthor{\bsnm{Scheck},~\bfnm{Adrienne~C.}\binits{A.~C.}},
  \bauthor{\bsnm{Tibshirani},~\bfnm{Robert}\binits{R.}} \AND
  \bauthor{\bsnm{Sikic},~\bfnm{Branimir~I.}\binits{B.~I.}}
(\byear{2009}).
\btitle{A network model of a cooperative genetic landscape in brain tumors}.
\bjournal{JAMA}
\bvolume{302}
\bpages{261--275}.
\bid{doi={10.1001/jama.2009.997}, issn={1538-3598}, pii={302/3/261},
  pmid={19602686}}
\bptok{imsref}%
\end{barticle}
\endbibitem

\bibitem[\protect\citeauthoryear{Cabanski et~al.}{2010}]{Cabanski}
\begin{barticle}[pbm]
\bauthor{\bsnm{Cabanski},~\bfnm{Christopher~R.}\binits{C.~R.}},
  \bauthor{\bsnm{Qi},~\bfnm{Yuan}\binits{Y.}},
  \bauthor{\bsnm{Yin},~\bfnm{Xiaoying}\binits{X.}},
  \bauthor{\bsnm{Bair},~\bfnm{Eric}\binits{E.}},
  \bauthor{\bsnm{Hayward},~\bfnm{Michele~C.}\binits{M.~C.}},
  \bauthor{\bsnm{Fan},~\bfnm{Cheng}\binits{C.}},
  \bauthor{\bsnm{Li},~\bfnm{Jianying}\binits{J.}},
  \bauthor{\bsnm{Wilkerson},~\bfnm{Matthew~D.}\binits{M.~D.}},
  \bauthor{\bsnm{Marron},~\bfnm{J.~S.}\binits{J.~S.}},
  \bauthor{\bsnm{Perou},~\bfnm{Charles~M.}\binits{C.~M.}} \AND
  \bauthor{\bsnm{Hayes},~\bfnm{D.~Neil}\binits{D.~N.}}
(\byear{2010}).
\btitle{SWISS MADE: Standardized within class sum of squares to evaluate
  methodologies and dataset elements}.
\bjournal{PLoS ONE}
\bvolume{5}
\bpages{e9905}.
\bid{doi={10.1371/journal.pone.0009905}, issn={1932-6203}, pmcid={2845619},
  pmid={20360852}}
\bptok{imsref}%
\end{barticle}
\endbibitem

\bibitem[\protect\citeauthoryear{Cancer Genome Atlas Research Network}{2008}]{TCGA}
\begin{bmisc}[auto]
\borganization{Cancer Genome Atlas Research Network}
(\byear{2008}).
\bhowpublished{Comprehensive genomic characterization defines human
  glioblastoma genes and core pathways. \textit{Nature} \textbf{455}
  1061--1068.}
\bid{doi={10.1038/nature07385}, issn={1476-4687}, mid={NIHMS68048},
  pii={nature07385}, pmcid={2671642}, pmid={18772890}}
\bptok{imsref}%
\end{bmisc}
\endbibitem

\bibitem[\protect\citeauthoryear{Candes et~al.}{2009}]{Candes}
\begin{bmisc}[author]
\bauthor{\bsnm{Candes},~\bfnm{E.}\binits{E.}},
  \bauthor{\bsnm{Li},~\bfnm{X.}\binits{X.}},
  \bauthor{\bsnm{Ma},~\bfnm{Y.}\binits{Y.}} \AND
  \bauthor{\bsnm{Wright},~\bfnm{J.}\binits{J.}}
(\byear{2009}).
\bhowpublished{Robust principal component analysis? Available at
  arXiv:\arxivurl{0912.3599}.}
\bptok{imsref}%
\end{bmisc}
\endbibitem

\bibitem[\protect\citeauthoryear{Di et~al.}{2009}]{Di}
\begin{barticle}[mr]
\bauthor{\bsnm{Di},~\bfnm{Chong-Zhi}\binits{C.-Z.}},
  \bauthor{\bsnm{Crainiceanu},~\bfnm{Ciprian~M.}\binits{C.~M.}},
  \bauthor{\bsnm{Caffo},~\bfnm{Brian~S.}\binits{B.~S.}} \AND
  \bauthor{\bsnm{Punjabi},~\bfnm{Naresh~M.}\binits{N.~M.}}
(\byear{2009}).
\btitle{Multilevel functional principal component analysis}.
\bjournal{Ann. Appl. Stat.}
\bvolume{3}
\bpages{458--488}.
\bid{doi={10.1214/08-AOAS206}, issn={1932-6157}, mr={2668715}}
\bptok{imsref}%
\end{barticle}
\endbibitem

\bibitem[\protect\citeauthoryear{Dweep et~al.}{2011}]{Dweep}
\begin{barticle}[pbm]
\bauthor{\bsnm{Dweep},~\bfnm{Harsh}\binits{H.}},
  \bauthor{\bsnm{Sticht},~\bfnm{Carsten}\binits{C.}},
  \bauthor{\bsnm{Pandey},~\bfnm{Priyanka}\binits{P.}} \AND
  \bauthor{\bsnm{Gretz},~\bfnm{Norbert}\binits{N.}}
(\byear{2011}).
\btitle{miRWalk-database: Prediction of possible miRNA binding sites by
  ``walking'' the genes of three genomes}.
\bjournal{J.~Biomed. Inform.}
\bvolume{44}
\bpages{839--847}.
\bid{doi={10.1016/j.jbi.2011.05.002}, issn={1532-0480},
  pii={S1532-0464(11)00078-5}, pmid={21605702}}
\bptok{imsref}%
\end{barticle}
\endbibitem

\bibitem[\protect\citeauthoryear{Fowler et~al.}{2011}]{Fowler}
\begin{barticle}[author]
\bauthor{\bsnm{Fowler},~\bfnm{A.}\binits{A.}},
  \bauthor{\bsnm{Thompson},~\bfnm{D.}\binits{D.}},
  \bauthor{\bsnm{Giles},~\bfnm{K.}\binits{K.}},
  \bauthor{\bsnm{Maleki},~\bfnm{S.}\binits{S.}},
  \bauthor{\bsnm{Mreich},~\bfnm{E.}\binits{E.}},
  \bauthor{\bsnm{Wheeler},~\bfnm{H.}\binits{H.}},
  \bauthor{\bsnm{Leedman},~\bfnm{P.}\binits{P.}},
  \bauthor{\bsnm{Biggs},~\bfnm{M.}\binits{M.}},
  \bauthor{\bsnm{Cook},~\bfnm{R.}\binits{R.}},
  \bauthor{\bsnm{Little},~\bfnm{N.}\binits{N.}},
  \bauthor{\bsnm{Robinson},~\bfnm{B.}\binits{B.}} \AND
  \bauthor{\bsnm{McDonald},~\bfnm{K.}\binits{K.}}
(\byear{2011}).
\btitle{miR-124a is frequently down-regulated in glioblastoma and is involved
  in migration and invasion.}
\bjournal{European Journal of Cancer}
\bvolume{47}
\bpages{953--963}.
\bptok{imsref}%
\end{barticle}
\endbibitem

\bibitem[\protect\citeauthoryear{Galberin and Cochrane}{2011}]{Galberin}
\begin{barticle}[author]
\bauthor{\bsnm{Galberin},~\bfnm{M.}\binits{M.}} \AND
  \bauthor{\bsnm{Cochrane},~\bfnm{G.}\binits{G.}}
(\byear{2011}).
\btitle{The 2011 nucleic acids research database issue and the online molecular
  biology database collection.}
\bjournal{Nucleic Acids Res.}
\bvolume{39}
\bpages{D1--D6}.
\bptok{imsref}%
\end{barticle}
\endbibitem

\bibitem[\protect\citeauthoryear{Gilad, Rifkin and Pritchard}{2008}]{Gilad}
\begin{barticle}[pbm]
\bauthor{\bsnm{Gilad},~\bfnm{Yoav}\binits{Y.}},
  \bauthor{\bsnm{Rifkin},~\bfnm{Scott~A.}\binits{S.~A.}} \AND
  \bauthor{\bsnm{Pritchard},~\bfnm{Jonathan~K.}\binits{J.~K.}}
(\byear{2008}).
\btitle{Revealing the architecture of gene regulation: The promise of eQTL
  studies}.
\bjournal{Trends Genet.}
\bvolume{24}
\bpages{408--415}.
\bid{doi={10.1016/j.tig.2008.06.001}, issn={0168-9525}, mid={NIHMS71884},
  pii={S0168-9525(08)00177-7}, pmcid={2583071}, pmid={18597885}}
\bptok{imsref}%
\end{barticle}
\endbibitem

\bibitem[\protect\citeauthoryear{Gillan et~al.}{2002}]{Gillan}
\begin{barticle}[author]
\bauthor{\bsnm{Gillan},~\bfnm{L.}\binits{L.}},
  \bauthor{\bsnm{Matei},~\bfnm{D.}\binits{D.}},
  \bauthor{\bsnm{Fishman},~\bfnm{D.}\binits{D.}},
  \bauthor{\bsnm{Gerbin},~\bfnm{C.}\binits{C.}},
  \bauthor{\bsnm{Karlan},~\bfnm{B.}\binits{B.}} \AND
  \bauthor{\bsnm{Chang},~\bfnm{D.}\binits{D.}}
(\byear{2002}).
\btitle{Periostin secreted by epithelial ovarian carcinoma is a ligand for
  alpha(V)beta(3) and alpha(V)beta(5) integrins and promotes cell motility.}
\bjournal{Cancer Research}
\bvolume{62}
\bpages{5358--5364}.
\bptok{imsref}%
\end{barticle}
\endbibitem

\bibitem[\protect\citeauthoryear{Hotelling}{1936}]{Hotelling}
\begin{barticle}[author]
\bauthor{\bsnm{Hotelling},~\bfnm{H.}\binits{H.}}
(\byear{1936}).
\btitle{Relations between two sets of variates.}
\bjournal{Biometrika}
\bvolume{28}
\bpages{321--377}.
\bptok{imsref}%
\end{barticle}
\endbibitem

\bibitem[\protect\citeauthoryear{L{\^e}~Cao et~al.}{2008}]{LeCao}
\begin{barticle}[mr]
\bauthor{\bsnm{L{\^e}~Cao},~\bfnm{Kim-Anh}\binits{K.-A.}},
  \bauthor{\bsnm{Rossouw},~\bfnm{Debra}\binits{D.}},
  \bauthor{\bsnm{Robert-Grani{\'e}},~\bfnm{Christ{\`e}le}\binits{C.}} \AND
  \bauthor{\bsnm{Besse},~\bfnm{Philippe}\binits{P.}}
(\byear{2008}).
\btitle{A sparse {PLS} for variable selection when integrating omics data}.
\bjournal{Stat. Appl. Genet. Mol. Biol.}
\bvolume{7}
\bpages{Art. 35, 31}.
\bid{doi={10.2202/1544-6115.1390}, issn={1544-6115}, mr={2457048}}
\bptok{imsref}%
\end{barticle}
\endbibitem

\bibitem[\protect\citeauthoryear{Lee et~al.}{2010}]{Lee}
\begin{barticle}[mr]
\bauthor{\bsnm{Lee},~\bfnm{Mihee}\binits{M.}},
  \bauthor{\bsnm{Shen},~\bfnm{Haipeng}\binits{H.}},
  \bauthor{\bsnm{Huang},~\bfnm{Jianhua~Z.}\binits{J.~Z.}} \AND
  \bauthor{\bsnm{Marron},~\bfnm{J.~S.}\binits{J.~S.}}
(\byear{2010}).
\btitle{Biclustering via sparse singular value decomposition}.
\bjournal{Biometrics}
\bvolume{66}
\bpages{1087--1095}.
\bid{doi={10.1111/j.1541-0420.2010.01392.x}, issn={0006-341X}, mr={2758496}}
\bptok{imsref}%
\end{barticle}
\endbibitem

\bibitem[\protect\citeauthoryear{Lock et~al.}{2012}]{Lock}
\begin{bmisc}[author]
\bauthor{\bsnm{Lock},~\bfnm{E.}\binits{E.}},
  \bauthor{\bsnm{Hoadley},~\bfnm{K.}\binits{K.}},
  \bauthor{\bsnm{Marron},~\bfnm{J.}\binits{J.}} \AND
  \bauthor{\bsnm{Nobel},~\bfnm{A.}\binits{A.}}
(\byear{2012}).
\bhowpublished{Supplement to ``Joint and individual variation explained (JIVE)
  for integrated analysis of multiple data types''.
  DOI:\doiurl{10.1214/12-AOAS597SUPP}.}
\bptok{imsref}%
\end{bmisc}
\endbibitem

\bibitem[\protect\citeauthoryear{Parkhomenko, Tritchler and
  Beyene}{2009}]{Parkhomenko}
\begin{barticle}[mr]
\bauthor{\bsnm{Parkhomenko},~\bfnm{Elena}\binits{E.}},
  \bauthor{\bsnm{Tritchler},~\bfnm{David}\binits{D.}} \AND
  \bauthor{\bsnm{Beyene},~\bfnm{Joseph}\binits{J.}}
(\byear{2009}).
\btitle{Sparse canonical correlation analysis with application to genomic data
  integration}.
\bjournal{Stat. Appl. Genet. Mol. Biol.}
\bvolume{8}
\bpages{Art. 1, 36}.
\bid{doi={10.2202/1544-6115.1406}, issn={1544-6115}, mr={2471148}}
\bptok{imsref}%
\end{barticle}
\endbibitem

\bibitem[\protect\citeauthoryear{Parkinson et~al.}{2009}]{Parkinson}
\begin{barticle}[author]
\bauthor{\bsnm{Parkinson},~\bfnm{H.}\binits{H.}},
  \bauthor{\bsnm{Kapushesky},~\bfnm{M.}\binits{M.}},
  \bauthor{\bsnm{Kolesnikov},~\bfnm{N.}\binits{N.}},
  \bauthor{\bsnm{Rustici},~\bfnm{G.}\binits{G.}},
  \bauthor{\bsnm{Shojatalab},~\bfnm{M.}\binits{M.}},
  \bauthor{\bsnm{Abeygunawardena},~\bfnm{N.}\binits{N.}},
  \bauthor{\bsnm{Berube},~\bfnm{H.}\binits{H.}},
  \bauthor{\bsnm{Dylag},~\bfnm{M.}\binits{M.}},
  \bauthor{\bsnm{Emam},~\bfnm{I.}\binits{I.}},
  \bauthor{\bsnm{Farne},~\bfnm{A.}\binits{A.}},
  \bauthor{\bsnm{Holloway},~\bfnm{E.}\binits{E.}},
  \bauthor{\bsnm{Lukk},~\bfnm{M.}\binits{M.}},
  \bauthor{\bsnm{Malone},~\bfnm{J.}\binits{J.}},
  \bauthor{\bsnm{Mani},~\bfnm{R.}\binits{R.}},
  \bauthor{\bsnm{Pilicheva},~\bfnm{E.}\binits{E.}},
  \bauthor{\bsnm{Rayner},~\bfnm{T.}\binits{T.}},
  \bauthor{\bsnm{Rezwan},~\bfnm{F.}\binits{F.}},
  \bauthor{\bsnm{Sharma},~\bfnm{A.}\binits{A.}},
  \bauthor{\bsnm{Williams},~\bfnm{E.}\binits{E.}},
  \bauthor{\bsnm{Bradley},~\bfnm{X.}\binits{X.}},
  \bauthor{\bsnm{Adamusiak},~\bfnm{T.}\binits{T.}},
  \bauthor{\bsnm{Brandizi},~\bfnm{M.}\binits{M.}},
  \bauthor{\bsnm{Burdett},~\bfnm{T.}\binits{T.}},
  \bauthor{\bsnm{Coulson},~\bfnm{R.}\binits{R.}},
  \bauthor{\bsnm{Krestyaninova},~\bfnm{M.}\binits{M.}},
  \bauthor{\bsnm{Kurnosov},~\bfnm{P.}\binits{P.}},
  \bauthor{\bsnm{Maguire},~\bfnm{E.}\binits{E.}},
  \bauthor{\bsnm{Neogi},~\bfnm{S.}\binits{S.}},
  \bauthor{\bsnm{Rocca-Serra},~\bfnm{P.}\binits{P.}},
  \bauthor{\bsnm{Sansone},~\bfnm{S.}\binits{S.}},
  \bauthor{\bsnm{Sklyar},~\bfnm{N.}\binits{N.}},
  \bauthor{\bsnm{Zhao},~\bfnm{M.}\binits{M.}},
  \bauthor{\bsnm{Sarkans},~\bfnm{U.}\binits{U.}} \AND
  \bauthor{\bsnm{Brazma},~\bfnm{A.}\binits{A.}}
(\byear{2009}).
\btitle{ArrayExpress update---from an archive of functional genomics experiments
  to the atlas of gene expression.}
\bjournal{Nucleic Acids Res.}
\bvolume{37}
\bpages{868--872}.
\bptok{imsref}%
\end{barticle}
\endbibitem

\bibitem[\protect\citeauthoryear{Peter}{2010}]{Peter}
\begin{barticle}[pbm]
\bauthor{\bsnm{Peter},~\bfnm{M.~E.}\binits{M.~E.}}
(\byear{2010}).
\btitle{Targeting of mRNAs by multiple miRNAs: The next step}.
\bjournal{Oncogene}
\bvolume{29}
\bpages{2161--2164}.
\bid{doi={10.1038/onc.2010.59}, issn={1476-5594}, pii={onc201059},
  pmid={20190803}}
\bptok{imsref}%
\end{barticle}
\endbibitem

\bibitem[\protect\citeauthoryear{Rhead et~al.}{2010}]{Rhead}
\begin{barticle}[author]
\bauthor{\bsnm{Rhead},~\bfnm{B.}\binits{B.}},
  \bauthor{\bsnm{Karolchik},~\bfnm{D.}\binits{D.}},
  \bauthor{\bsnm{Kuhn},~\bfnm{R.}\binits{R.}},
  \bauthor{\bsnm{Hinrichs},~\bfnm{A.}\binits{A.}},
  \bauthor{\bsnm{Zweig},~\bfnm{A.}\binits{A.}},
  \bauthor{\bsnm{Fujita},~\bfnm{P.}\binits{P.}},
  \bauthor{\bsnm{Diekhans},~\bfnm{M.}\binits{M.}},
  \bauthor{\bsnm{Smith},~\bfnm{K.}\binits{K.}},
  \bauthor{\bsnm{Rosenbloom},~\bfnm{K.}\binits{K.}},
  \bauthor{\bsnm{Raney},~\bfnm{B.}\binits{B.}},
  \bauthor{\bsnm{Pohl},~\bfnm{A.}\binits{A.}},
  \bauthor{\bsnm{Pheasant},~\bfnm{M.}\binits{M.}},
  \bauthor{\bsnm{Meyer},~\bfnm{L.}\binits{L.}},
  \bauthor{\bsnm{Learned},~\bfnm{K.}\binits{K.}},
  \bauthor{\bsnm{Hsu},~\bfnm{F.}\binits{F.}},
  \bauthor{\bsnm{Hillman-Jackson},~\bfnm{J.}\binits{J.}},
  \bauthor{\bsnm{Harte},~\bfnm{R.}\binits{R.}},
  \bauthor{\bsnm{Giardine},~\bfnm{B.}\binits{B.}},
  \bauthor{\bsnm{Dreszer},~\bfnm{T.}\binits{T.}},
  \bauthor{\bsnm{Clawson},~\bfnm{H.}\binits{H.}},
  \bauthor{\bsnm{Barber},~\bfnm{G.}\binits{G.}},
  \bauthor{\bsnm{Haussler},~\bfnm{D.}\binits{D.}} \AND
  \bauthor{\bsnm{Kent},~\bfnm{W.}\binits{W.}}
(\byear{2010}).
\btitle{The {U}{C}{S}{C} genome browser database: Update 2010.}
\bjournal{Nucleic Acids Res.}
\bvolume{38}
\bpages{613--619}.
\bptok{imsref}%
\end{barticle}
\endbibitem

\bibitem[\protect\citeauthoryear{Schwarz}{1978}]{Schwarz}
\begin{barticle}[mr]
\bauthor{\bsnm{Schwarz},~\bfnm{Gideon}\binits{G.}}
(\byear{1978}).
\btitle{Estimating the dimension of a model}.
\bjournal{Ann. Statist.}
\bvolume{6}
\bpages{461--464}.
\bid{issn={0090-5364}, mr={0468014}}
\bptok{imsref}%
\end{barticle}
\endbibitem

\bibitem[\protect\citeauthoryear{Shen and Huang}{2008}]{ShenHuang}
\begin{barticle}[mr]
\bauthor{\bsnm{Shen},~\bfnm{Haipeng}\binits{H.}} \AND
  \bauthor{\bsnm{Huang},~\bfnm{Jianhua~Z.}\binits{J.~Z.}}
(\byear{2008}).
\btitle{Sparse principal component analysis via regularized low rank matrix
  approximation}.
\bjournal{J. Multivariate Anal.}
\bvolume{99}
\bpages{1015--1034}.
\bid{doi={10.1016/j.jmva.2007.06.007}, issn={0047-259X}, mr={2419336}}
\bptok{imsref}%
\end{barticle}
\endbibitem

\bibitem[\protect\citeauthoryear{Shen, Olshen and Ladanyi}{2009}]{icluster}
\begin{barticle}[pbm]
\bauthor{\bsnm{Shen},~\bfnm{Ronglai}\binits{R.}},
  \bauthor{\bsnm{Olshen},~\bfnm{Adam~B.}\binits{A.~B.}} \AND
  \bauthor{\bsnm{Ladanyi},~\bfnm{Marc}\binits{M.}}
(\byear{2009}).
\btitle{Integrative clustering of multiple genomic data types using a joint
  latent variable model with application to breast and lung cancer subtype
  analysis}.
\bjournal{Bioinformatics}
\bvolume{25}
\bpages{2906--2912}.
\bid{doi={10.1093/bioinformatics/btp543}, issn={1367-4811}, pii={btp543},
  pmcid={2800366}, pmid={19759197}}
\bptok{imsref}%
\end{barticle}
\endbibitem

\bibitem[\protect\citeauthoryear{Sporns, Tononi and
  K{\"{o}}tter}{2005}]{Sporns}
\begin{barticle}[pbm]
\bauthor{\bsnm{Sporns},~\bfnm{Olaf}\binits{O.}},
  \bauthor{\bsnm{Tononi},~\bfnm{Giulio}\binits{G.}} \AND
  \bauthor{\bsnm{K{\"{o}}tter},~\bfnm{Rolf}\binits{R.}}
(\byear{2005}).
\btitle{The human connectome: A structural description of the human brain}.
\bjournal{PLoS Comput. Biol.}
\bvolume{1}
\bpages{e42}.
\bid{doi={10.1371/journal.pcbi.0010042}, issn={1553-7358}, pmcid={1239902},
  pmid={16201007}}
\bptok{imsref}%
\end{barticle}
\endbibitem

\bibitem[\protect\citeauthoryear{Tibshirani}{1996}]{Tibshirani}
\begin{barticle}[mr]
\bauthor{\bsnm{Tibshirani},~\bfnm{Robert}\binits{R.}}
(\byear{1996}).
\btitle{Regression shrinkage and selection via the lasso}.
\bjournal{J. Roy. Statist. Soc. Ser. B}
\bvolume{58}
\bpages{267--288}.
\bid{issn={0035-9246}, mr={1379242}}
\bptok{imsref}%
\end{barticle}
\endbibitem

\bibitem[\protect\citeauthoryear{Trygg and Wold}{2003}]{Trygg}
\begin{barticle}[author]
\bauthor{\bsnm{Trygg},~\bfnm{J.}\binits{J.}} \AND
  \bauthor{\bsnm{Wold},~\bfnm{S.}\binits{S.}}
(\byear{2003}).
\btitle{O2-{P}{L}{S}, a two-block ({X}-{Y}) latent variable regression
  ({L}{V}{R}) method with an integral {O}{S}{C} filter.}
\bjournal{Journal of Chemometrics}
\bvolume{17}
\bpages{53--64}.
\bptok{imsref}%
\end{barticle}
\endbibitem

\bibitem[\protect\citeauthoryear{Verhaak et~al.}{2010}]{Verheek}
\begin{barticle}[pbm]
\bauthor{\bsnm{Verhaak},~\bfnm{Roel G.~W.}\binits{R.~G.~W.}},
  \bauthor{\bsnm{Hoadley},~\bfnm{Katherine~A.}\binits{K.~A.}},
  \bauthor{\bsnm{Purdom},~\bfnm{Elizabeth}\binits{E.}},
  \bauthor{\bsnm{Wang},~\bfnm{Victoria}\binits{V.}},
  \bauthor{\bsnm{Qi},~\bfnm{Yuan}\binits{Y.}},
  \bauthor{\bsnm{Wilkerson},~\bfnm{Matthew~D.}\binits{M.~D.}},
  \bauthor{\bsnm{Miller},~\bfnm{C.~Ryan}\binits{C.~R.}},
  \bauthor{\bsnm{Ding},~\bfnm{Li}\binits{L.}},
  \bauthor{\bsnm{Golub},~\bfnm{Todd}\binits{T.}},
  \bauthor{\bsnm{Mesirov},~\bfnm{Jill~P.}\binits{J.~P.}},
  \bauthor{\bsnm{Alexe},~\bfnm{Gabriele}\binits{G.}},
  \bauthor{\bsnm{Lawrence},~\bfnm{Michael}\binits{M.}},
  \bauthor{\bsnm{O'Kelly},~\bfnm{Michael}\binits{M.}},
  \bauthor{\bsnm{Tamayo},~\bfnm{Pablo}\binits{P.}},
  \bauthor{\bsnm{Weir},~\bfnm{Barbara~A.}\binits{B.~A.}},
  \bauthor{\bsnm{Gabriel},~\bfnm{Stacey}\binits{S.}},
  \bauthor{\bsnm{Winckler},~\bfnm{Wendy}\binits{W.}},
  \bauthor{\bsnm{Gupta},~\bfnm{Supriya}\binits{S.}},
  \bauthor{\bsnm{Jakkula},~\bfnm{Lakshmi}\binits{L.}},
  \bauthor{\bsnm{Feiler},~\bfnm{Heidi~S.}\binits{H.~S.}},
  \bauthor{\bsnm{Hodgson},~\bfnm{J.~Graeme}\binits{J.~G.}},
  \bauthor{\bsnm{James},~\bfnm{C.~David}\binits{C.~D.}},
  \bauthor{\bsnm{Sarkaria},~\bfnm{Jann~N.}\binits{J.~N.}},
  \bauthor{\bsnm{Brennan},~\bfnm{Cameron}\binits{C.}},
  \bauthor{\bsnm{Kahn},~\bfnm{Ari}\binits{A.}},
  \bauthor{\bsnm{Spellman},~\bfnm{Paul~T.}\binits{P.~T.}},
  \bauthor{\bsnm{Wilson},~\bfnm{Richard~K.}\binits{R.~K.}},
  \bauthor{\bsnm{Speed},~\bfnm{Terence~P.}\binits{T.~P.}},
  \bauthor{\bsnm{Gray},~\bfnm{Joe~W.}\binits{J.~W.}},
  \bauthor{\bsnm{Meyerson},~\bfnm{Matthew}\binits{M.}},
  \bauthor{\bsnm{Getz},~\bfnm{Gad}\binits{G.}},
  \bauthor{\bsnm{Perou},~\bfnm{Charles~M.}\binits{C.~M.}},
  \bauthor{\bsnm{Hayes},~\bfnm{D.~Neil}\binits{D.~N.}} \AND
  \bauthor{\bsnm{{Cancer Genome Atlas Research Network}}}
(\byear{2010}).
\btitle{Integrated genomic analysis identifies clinically relevant subtypes of
  glioblastoma characterized by abnormalities in PDGFRA, IDH1, EGFR, and NF1}.
\bjournal{Cancer Cell}
\bvolume{17}
\bpages{98--110}.
\bid{doi={10.1016/j.ccr.2009.12.020}, issn={1878-3686}, mid={NIHMS166306},
  pii={S1535-6108(09)00432-2}, pmcid={2818769}, pmid={20129251}}
\bptok{imsref}%
\end{barticle}
\endbibitem

\bibitem[\protect\citeauthoryear{Westerhuis, Kourti and
  MacGregor}{1998}]{Westerhuis}
\begin{barticle}[author]
\bauthor{\bsnm{Westerhuis},~\bfnm{J.}\binits{J.}},
  \bauthor{\bsnm{Kourti},~\bfnm{T.}\binits{T.}} \AND
  \bauthor{\bsnm{MacGregor},~\bfnm{J.}\binits{J.}}
(\byear{1998}).
\btitle{Analysis of multiblock and hierarchical PCA and PLS models.}
\bjournal{Journal of Chemometrics}
\bvolume{12}
\bpages{301--321}.
\bptok{imsref}%
\end{barticle}
\endbibitem

\bibitem[\protect\citeauthoryear{Witten and Tibshirani}{2009}]{Witten}
\begin{barticle}[mr]
\bauthor{\bsnm{Witten},~\bfnm{Daniela~M.}\binits{D.~M.}} \AND
  \bauthor{\bsnm{Tibshirani},~\bfnm{Robert~J.}\binits{R.~J.}}
(\byear{2009}).
\btitle{Extensions of sparse canonical correlation analysis with applications
  to genomic data}.
\bjournal{Stat. Appl. Genet. Mol. Biol.}
\bvolume{8}
\bpages{Art. 28, 29}.
\bid{doi={10.2202/1544-6115.1470}, issn={1544-6115}, mr={2533636}}
\bptok{imsref}%
\end{barticle}
\endbibitem

\bibitem[\protect\citeauthoryear{Wold}{1985}]{Hwold}
\begin{bincollection}[author]
\bauthor{\bsnm{Wold},~\bfnm{H.}\binits{H.}}
(\byear{1985}).
\btitle{Partial Least Squares}.
In \bbooktitle{Encyclopedia of Statistical Sciences (Vol. 6)}
(\beditor{\bfnm{S.}\binits{S.}~\bsnm{Kotz}} \AND
  \beditor{\bfnm{N.}\binits{N.}~\bsnm{Johnson}}, eds.)
\bpages{581--591}.
\bpublisher{Wiley}, \blocation{New York}.
\bptok{imsref}%
\end{bincollection}
\endbibitem

\bibitem[\protect\citeauthoryear{Wold, Kettaneh and Tjessem}{1996}]{Wold}
\begin{barticle}[author]
\bauthor{\bsnm{Wold},~\bfnm{S.}\binits{S.}},
  \bauthor{\bsnm{Kettaneh},~\bfnm{N.}\binits{N.}} \AND
  \bauthor{\bsnm{Tjessem},~\bfnm{K.}\binits{K.}}
(\byear{1996}).
\btitle{Hierarchical multiblock {P}{L}{S} and {P}{C} models for easier model
  interpretation and as an alternative to variable selection.}
\bjournal{Journal of Chemometrics}
\bvolume{10}
\bpages{463--482}.
\bptok{imsref}%
\end{barticle}
\endbibitem

\bibitem[\protect\citeauthoryear{Zinn et~al.}{2011}]{Zinn}
\begin{barticle}[author]
\bauthor{\bsnm{Zinn},~\bfnm{P.}\binits{P.}},
  \bauthor{\bsnm{Majadan},~\bfnm{B.}\binits{B.}},
  \bauthor{\bsnm{Sathyan},~\bfnm{P.}\binits{P.}},
  \bauthor{\bsnm{Singh},~\bfnm{K.}\binits{K.}},
  \bauthor{\bsnm{Majumder},~\bfnm{S.}\binits{S.}},
  \bauthor{\bsnm{Jolesz},~\bfnm{F.}\binits{F.}} \AND
  \bauthor{\bsnm{Colen},~\bfnm{R.}\binits{R.}}
(\byear{2011}).
\btitle{Radiogenomic mapping of edema/cellular invasion {M}{R}{I}-phenotypes in
  glioblastoma multiforme.}
\bjournal{PLoS ONE}
\bvolume{6}
\bpages{e25451}.
\bptok{imsref}%
\end{barticle}
\endbibitem

\end{thebibliography}
\end{document}